\documentclass[sigconf]{acmart}
\renewcommand\footnotetextcopyrightpermission[1]{}
\usepackage{comment}
\usepackage[normalem]{ulem}
\usepackage{subcaption}
\AtBeginDocument{%
  }

\setcopyright{acmlicensed}
\copyrightyear{2026}
\acmYear{2026}
\acmDOI{XXXXXXX.XXXXXXX}
\acmConference[ACM 2026]{Make sure to enter the correct conference title from your rights confirmation email}{Feb 1, 2026}{USA}
\acmISBN{978-1-4503-XXXX-X/2018/06}

\newcommand{\tsnote}[1]{\textcolor{blue}{#1}}
\renewcommand{\tsnote}[1]{}

\newcommand{\squishlist}{
\begin{list}{$\bullet$}
  { \setlength{\itemsep}{0pt}
     \setlength{\parsep}{0pt}
     \setlength{\topsep}{0pt}
     \setlength{\partopsep}{0pt}
     \setlength{\leftmargin}{0em}
     \setlength{\labelwidth}{0em}
     \setlength{\labelsep}{0.2em} } }

\newcommand{\squishlisttwo}{
\begin{list}{$\bullet$}
  { \setlength{\itemsep}{0pt}
     \setlength{\parsep}{0pt}
    \setlength{\topsep}{0pt}
    \setlength{\partopsep}{0pt}
    \setlength{\leftmargin}{2em}
    \setlength{\labelwidth}{1.5em}
    \setlength{\labelsep}{0.5em} } }

\newcommand{\squishend}{
  \end{list}  }

\begin{document}


\title{From Feasibility to Desirability: Plan, Learn, Adapt (PLA) Framework for Personalized On-Device Itinerary Generation}



\author{Himel Dev\textsuperscript{1}, Tanmoy Sen\textsuperscript{2}, Madhusudan Basak\textsuperscript{3}, Bashima Islam\textsuperscript{3}}
\affiliation{%
  \institution{%
    \textsuperscript{1}529 Tech LLC, \textsuperscript{2}University of Virginia, \textsuperscript{3}University of Massachusetts Amherst\\
    himel@529-tech.com, ts5xm@virginia.edu, mbasak@umass.edu, bashima@umass.edu
  }
  \country{USA}
}

\begin{abstract}
Generating personalized trip itineraries is a complex planning task and involves a tension between hard combinatorial feasibility and soft latent desirability. Classical optimization enforces constraints but fails to capture subjective traveler preferences. While learning-based approaches model preferences, they cannot guarantee feasibility. Mobile deployment imposes additional resource constraints on both. To address this, we propose Plan, Learn, Adapt (PLA), a three-stage framework for personalized on-device itinerary generation. The \textit{Plan} stage builds a heterogeneous ensemble of lightweight planners that produces structurally diverse feasible candidates. From pairwise itinerary comparisons, \textit{Learn} fits a compact Bradley-Terry reward model that captures emergent schedule properties such as pacing, geographic coherence, and day balance, which per-POI signals miss. Finally,  \textit{Adapt} applies feasibility-preserving local refinement within a device-aware compute budget; every intermediate state is feasible by construction.
On 2,519 pairwise human comparisons across 100+ U.S. cities, the reward-guided ensemble achieves a 67.8\% win rate (+11.2 points over the best single planner) with 100\% feasibility, while three frontier LLMs (GPT-5, Claude Opus 4.5, Gemini 3 Pro) achieve 0\% feasibility on the same constraints. The reward model generalizes across held-out cities, with a 67.6\% mean leave-one-city-out accuracy. In production deployment within FlyEnJoy, PLA increased itinerary completion rates by 91\% with 109.9 ms average on-device latency.

\end{abstract}


\maketitle

\section{Introduction}
Multi-day trip itinerary generation is a canonical human-aligned combinatorial planning problem. The system must select, order, and time a set of points of interest (POIs) across multiple days under hard spatio-temporal constraints (opening hours, availability windows, daily time budgets) while also satisfying soft desirability criteria such as pacing, diversity, and alignment with stated interests.

\begin{figure}[t]
    \centering
    \includegraphics[width=0.9\linewidth]{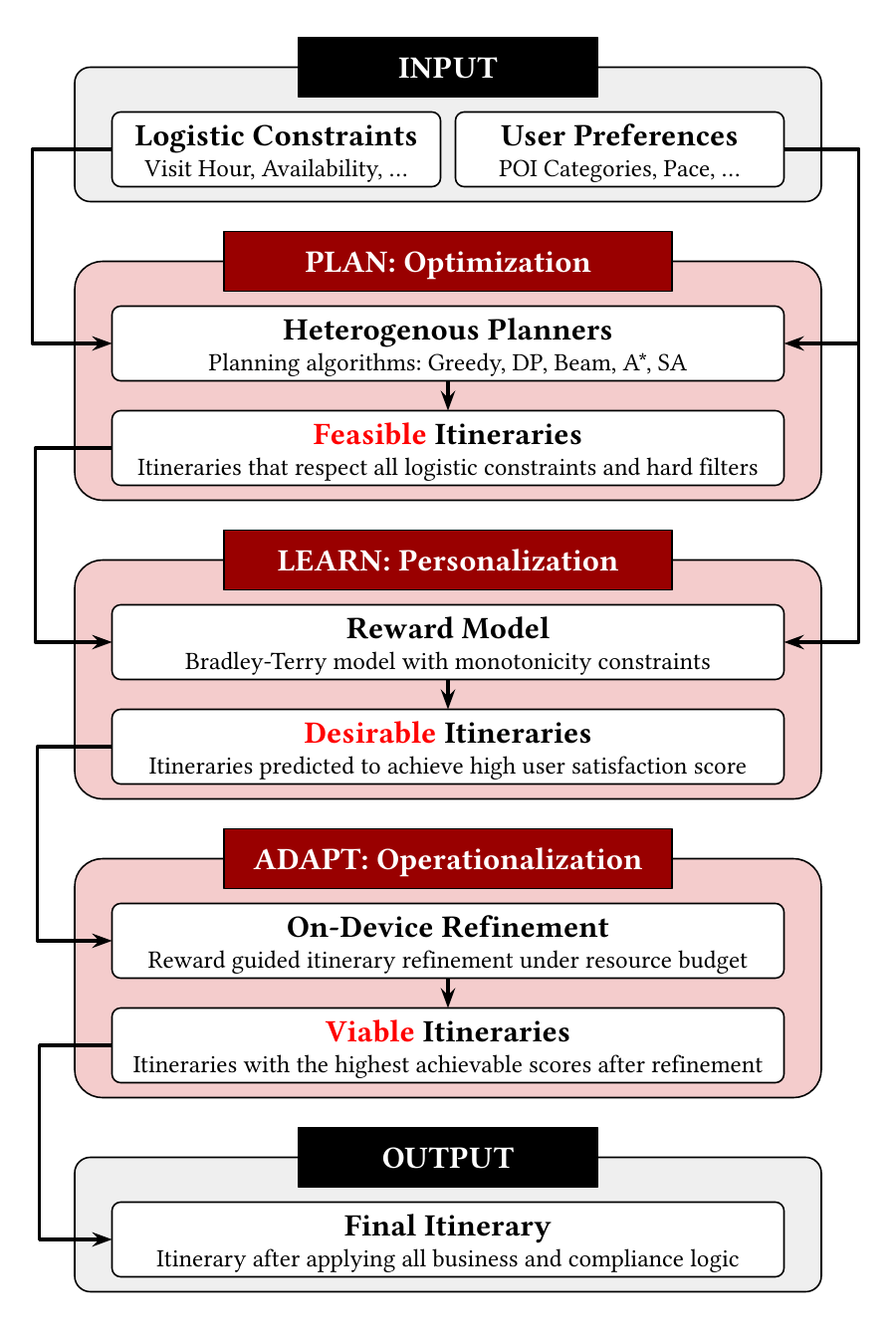}\vspace{-0.2in}
    \caption{Overview of the PLA framework. \textsc{Plan} produces a diverse set of feasible seed itineraries; \textsc{Learn} fits a human-aligned reward model from pairwise preferences; \textsc{Adapt} then refines the selected itinerary on-device under a resource budget while preserving feasibility throughout.
    \vspace{-0.2in}}
    \label{fig:system_overview}
\end{figure}

This duality creates a tension between feasibility and desirability. Classical planners~\cite{lim2015personalized,zhang2016trip} enforce constraints reliably but rely on hand-crafted objectives that are weak proxies for human satisfaction. In our earlier deployment, itineraries with similar heuristic scores were often judged very differently by users, reflecting preferences that no single composite score captures. Learning-based approaches~\cite{chen2020itinerary,chen2022automatic} capture preferences but, without explicit constraint handling, frequently produce invalid schedules. LLM-based approaches~\cite{li2024large,volchek2024chatgpt} face the same limitation. In our benchmark (Section~\ref{sec:e2e_evaluation}), none of the 294 itineraries from GPT-5, Claude Opus 4.5, and Gemini 3 Pro satisfied all hard constraints simultaneously.


A third requirement is viability. Solutions must be deployable at scale on end-user devices. Our product, \textbf{\textsc{FlyEnJoy}}\footnote{\url{https://apps.apple.com/app/6745104853}}, is an offline-first iOS travel app whose users often need itineraries in low or no connectivity settings (airports, in-flight, international roaming) with strict latency expectations. Cloud-dependent LLM and learning-based approaches are costly at scale and incompatible with offline, latency-sensitive, privacy-preserving use.

These three requirements, feasibility, desirability, and viability, are unlikely to be met by any single paradigm. We propose \textsc{Plan, Learn, Adapt (PLA)}, a three-stage framework for personalized on-device itinerary generation (Figure~\ref{fig:system_overview}). In Section~\ref{sec:seed_algorithms}, the \textsc{Plan} stage produces structurally diverse feasible candidates through a heterogeneous planner ensemble. Section~\ref{sec:learn} then describes how \textsc{Learn} fits a compact reward model from pairwise itinerary comparisons. Section~\ref{sec:adapt} introduces \textsc{Adapt}, which carries out reward-guided local refinement under a device-aware compute budget while preserving feasibility.

Our main contributions are as follows.
\squishlist
    \item \textbf{Heterogeneous Planner Ensemble:}
   No single planning algorithm dominates across trip contexts; we establish this empirically rather than assume it. Reward-guided selection over a heterogeneous planner ensemble yields a 67.8\% win rate, +11.2 pp over the best individual planner (Section~\ref{sec:seed_algorithms}).
    \item \textbf{Itinerary-Level Preference Modeling:}
    Itinerary quality involves emergent properties (pacing, geographic coherence, day balance) that per-POI scores cannot capture. We collect 2,519 pairwise comparisons of complete multi-day itineraries across 100+ U.S. cities and train a compact Bradley–Terry reward model achieving 67.6\% mean leave-one-city-out accuracy (Section ~\ref{sec:learn}).
    \item \textbf{On-Device Itinerary Refinement:}
    \textsc{Adapt} improves desirability via reward-guided local edits while guaranteeing every intermediate state is feasible by construction (not merely the final output). A device-aware budget keeps on-device latency to 109.9 ms on average, enabling fully offline, interactive-speed execution (Section~\ref{sec:adapt}).
\squishend

To support reproducibility, we publicly release a Python implementation of the core PLA components along with sample data.\footnote{\url{https://github.com/Official529Tech/pla-itinerary}}

\section{System Overview}

\noindent \textbf{Problem Setup.} Given a trip context (including destination, duration, availability, budget, interests, etc.) and a set of candidate POIs (with locations, opening hours, and visit durations), the goal is to generate a multi-day itinerary that is both feasible and desirable. Feasibility imposes opening-hour, availability, travel-time, and time-budget constraints (Section~\ref{sec:seed_algorithms}). Desirability reflects latent user satisfaction at the itinerary level and is not directly observable. We learn a compact surrogate reward from pairwise human preferences over complete itineraries (Section~\ref{sec:learn}), optimized under a device-aware compute budget (Section~\ref{sec:adapt}).

\subsection{PLAN: Diverse Feasible Itinerary Generation}
\label{sec:seed_algorithms}

We implement five heterogeneous planners for multi-day itinerary generation under a shared feasibility layer. Each planner outputs a feasible itinerary (an ordered, time-stamped sequence of POI visits) that respects all hard constraints and optimizes a shared composite objective. Although the objective is shared, the planners' distinct search paradigms yield \emph{structurally diverse} candidates differing in POI selection, ordering, and time allocation. The planner suite is designed to supply a diverse candidate pool for downstream selection and refinement, since identifying a single best planner across all trip contexts is not feasible.

\subsubsection{POI Data and User Inputs}
\label{sec:inputs}
Our system operates over city-specific POI datasets covering 100+ U.S. cities. Each POI $p$ has opening hours, visit duration, cost tier, location (with a precomputed travel-time zone), categories, and a popularity score derived from user ratings. A trip context $u$ specifies destination, duration, categories of interest, budget tier, availability windows, and pace (Relaxed, Balanced, or Packed), which sets the daily time budget. Together, these inputs define the feasible search space while remaining minimal enough for fast, on-device planning.

\subsubsection{Feasibility Constraints}
\label{sec:constraints_shared}

All planners enforce the following hard constraints that define a valid, executable itinerary:
\squishlist
    \item \textbf{Visit Hours:} Each visit must start after opening and end before closing, i.e., $\mathit{start}(p) \ge \mathit{open}(p)$ and $\mathit{start}(p) + \mathit{dur}(p) \le \mathit{close}(p)$.
    \item \textbf{Availability Windows:} Every visit must lie entirely within the traveler's available time on that day.
    \item \textbf{Travel-Time Consistency:} Sufficient time must exist between consecutive visits to travel between them. If arrival precedes opening, the traveler waits.
    \item \textbf{Time Budget:} Total active time (visits plus travel) on each day must not exceed the budget implied by the selected pace.
    \item \textbf{Cost Budget:} POIs with $\mathit{cost}(p)$ exceeding the user's budget tier are excluded from consideration.
    \item \textbf{No Repeats:} Each POI is visited at most once per trip.
\squishend

All planners share a \emph{feasibility layer} that computes earliest-feasible arrival times (with waiting when arrival precedes opening) and enforces day- and trip-level constraints uniformly, so cross-planner differences reflect search strategy rather than inconsistent constraint handling.

\subsubsection{Objective Components}
\label{sec:objective_components}
Beyond feasibility, planners optimize for itinerary quality using a hand-crafted, interpretable objective. We score a candidate POI $p$ at its earliest feasible start time $t_s$, given the current time cursor $t$ and previously scheduled POI $p_\ell$:

\squishlist
    \item \textbf{Travel Efficiency:}
We penalize long travel times to encourage geographically coherent itineraries:
$\mathit{travel}(p_\ell, p) = T(\mathit{zone}(p_\ell), \mathit{zone}(p))$.

\item \textbf{Waiting Time:}
Arriving before a POI opens incurs idle time, which we penalize:
$\mathit{wait}(p, t) = \max\{0,\; \mathit{open}(p) - (t + \mathit{travel}(p_\ell, p))\}$.

\item \textbf{Scheduling Urgency:}
POIs with narrow remaining windows should be prioritized to avoid missing them entirely. We define urgency based on the gap between visit end and closing time:
$\mathit{urgency}(p, t_s) = \frac{1}{\varepsilon + \max\{0,\; \mathit{close}(p) - (t_s + \mathit{dur}(p))\}}$,
where $\varepsilon$ is a small constant for numerical stability. POIs that would barely fit receive high urgency scores.

\item \textbf{Category Diversity:}
To avoid monotonous itineraries (e.g., five museums in a row), we reward visits that introduce under-represented categories at both day and trip levels:
$\mathit{div}(p) = \lambda_d \sum_{c \in \mathit{cat}(p)} \frac{\gamma_c}{1 + \mathit{count}_d(c)} + \lambda_t \sum_{c \in \mathit{cat}(p)} \frac{\gamma_c}{1 + \mathit{count}_t(c)}$,
where $\mathit{count}_d(c)$ and $\mathit{count}_t(c)$ are the number of visits to category $c$ on the current day and trip respectively, $\gamma_c$ is a category weight (higher for user-preferred categories), and $\lambda_d, \lambda_t$ control the day vs.\ trip diversity trade-off.

\item \textbf{POI Popularity:}
Higher-quality POIs, as measured by $\mathit{pop}(p)$, contribute more value to the itinerary.
\squishend

\subsubsection{Shared Composite Score}
\label{sec:composite_score}

The objective components are combined into a weighted composite score used by constructive planners (Greedy, Beam, A*) for step-level decisions:
\begin{equation}
\label{eq:composite_score}
\begin{aligned}
\mathit{score}(p, t_s) \;=\;&
w_{\text{pop}} \cdot \mathit{pop}(p)
- w_{\text{travel}} \cdot \mathit{travel}(p_\ell, p)
- w_{\text{wait}} \cdot \mathit{wait}(p, t) \\
&+ w_{\text{urg}} \cdot \mathit{urgency}(p, t_s)
+ w_{\text{div}} \cdot \mathit{div}(p).
\end{aligned}
\end{equation}

All weights $(w_{\text{pop}}, w_{\text{travel}}, w_{\text{wait}}, w_{\text{urg}}, w_{\text{div}})$ are configurable hyperparameters shared across planners for fair comparison. Global optimization planners and local search use analogous objectives adapted to their formulations. The shared objective structure ensures that performance differences across planners stem from search strategy rather than from inconsistent quality definitions. 

\subsubsection{Planning Algorithms}
\label{sec:planners}
We implement five planning algorithms across four paradigms: constructive heuristics (Greedy), dynamic programming (DP), tree search (Beam, A*), and local search (SA). Table~\ref{tab:planner-comparison} summarizes the key characteristics of each planning algorithm. All algorithms share the feasibility layer and composite objective (Eq.~\ref{eq:composite_score}), differing only in search strategy.


\squishlist
    \item \textbf{Greedy:} Iteratively schedules the highest-scoring feasible POI at its earliest valid start time.
    \item \textbf{Dynamic Programming (DP):} Resource-bounded optimization over a time-expanded state space maximizing POI popularity and coverage.
    \item \textbf{Beam Search:} Anytime tree search over a bounded frontier of partial itineraries, pruning by composite score.
    \item \textbf{A* Search:} Best-first search with an admissible upper bound on remaining reward.
    \item \textbf{Simulated Annealing (SA).} Local search with probabilistic acceptance of structural edits over a feasible seed.
\squishend

\begin{table}[t]
\centering
\small
\caption{Comparison of planning algorithms and trade-offs.\vspace{-0.05in}}
\label{tab:planner-comparison}
\resizebox{\linewidth}{!}{
\begin{tabular}{lll}
\toprule
\textbf{Algorithm} & \textbf{Paradigm} & \textbf{Key Trade-off} \\
\midrule
Greedy & Constructive & Solution quality vs.\ speed \\
DP & Dynamic prog. & Optimality vs.\ state space \\
Beam & Tree search & Solution quality vs.\ search breadth \\
A* & Tree search & Optimality vs.\ search cost \\
SA & Local search & Exploration vs.\ convergence speed \\
\bottomrule
\end{tabular}
}
\vspace{-0.2in}
\end{table}

We also prototyped a Constraint Programming (CP-SAT) planner but excluded it from deployment because Google OR-Tools lacks a native Swift implementation.


\subsection{LEARN: Human Preference Modeling}
\label{sec:learn}

User satisfaction with an itinerary is latent and context-dependent. We therefore conduct a pairwise preference study over complete itineraries and train a compact reward model over interpretable itinerary-level features.

\subsubsection{Preference Data Collection}
\label{sec:preference_collection}
We collect preferences via a web platform that presents side-by-side itinerary pairs from randomly chosen planners. Planner identities are hidden and left-right placement is randomized to prevent ordering bias. Guidelines emphasize holistic itinerary-level judgment; submissions with completion time well below median are filtered. This yields 2,519 pairwise comparisons across 100+ U.S. cities. We use a feature-based Bradley–Terry formulation for data efficiency: each comparison supervises ${\sim}20$ interpretable features simultaneously (Section~\ref{sec:reward_model}).

\subsubsection{Reward Model Architecture}
\label{sec:reward_model}
We learn a scoring function $R(x)$ that ranks itineraries by predicted human preference.

\noindent \textbf{Feature Engineering.} \label{sec:features} 
Each itinerary $x$ is mapped to a fixed dimensional feature vector $\phi(x) \in \mathbb{R}^d$, and we learn $R_\theta(x) = f_\theta(\phi(x))$. The feature set ($d \approx 20$) captures structural, temporal, and preference-aligned properties. Table~\ref{tab:features} summarizes the feature groups. All features are computed deterministically from POI metadata and the generated schedule (operating hours, visit durations, travel times, category counts) rather than from city-specific or cultural signals. The representation is therefore portable across destinations: the model learns that excessive travel time or imbalanced daily load reduce preference, properties that hold city to city. User-context interaction terms (e.g., \texttt{is\_packed\_x\_num\_pois}, \texttt{is\_\\relaxed\_x\_avg\_gap}) condition itinerary scores on the traveler's stated preferences. A POI-dense schedule then ranks higher for a packed traveler than a relaxed one, providing context-conditioned personalization without per-user history.

\begin{table}[t]
\centering
\small
\caption{Itinerary-level features for reward modeling.\vspace{-0.05in}}
\label{tab:features}
\resizebox{\linewidth}{!}{
\begin{tabular}{ll}
\toprule
\textbf{Category} & \textbf{Feature Example} \\
\midrule
POI Quality & Count, total/avg/min/max popularity \\
Category Diversity & Unique categories, entropy, alignment \\
Temporal Efficiency & Travel time, wait time, travel-to-activity ratio \\
Schedule Balance & Daily load, time utilization, max consecutive travel \\
Constraint Slack & Budget margin, closing-time slack \\
\bottomrule
\end{tabular}
}
\vspace{-0.15in}
\end{table}

\noindent \textbf{Learning Objective.}
We adopt the Bradley–Terry framework~\cite{bradley1952rank,sun2024rethinking}: the probability of preferring $x_a$ over $x_b$ follows a logistic model
$P(x_a \succ x_b) = \sigma(R_\theta(x_a) - R_\theta(x_b))$,
where $\sigma(\cdot)$ is the sigmoid. Given a dataset  $\mathcal{D} = \{(x_a^{(i)}, x_b^{(i)}, y^{(i)})\}_{i=1}^N$ where $y^{(i)} \in \{-1, +1\}$, we minimize the regularized negative log-likelihood and our loss is
$\mathcal{L}(\theta) = -\frac{1}{N}\sum_{i=1}^N \log \sigma\left(y^{(i)} \cdot (R_\theta(x_a^{(i)}) - R_\theta(x_b^{(i)}))\right) + \lambda \|\theta\|_2^2$. Here, $\lambda$ controls $\ell_2$ regularization to prevent overfitting.   



\noindent \textbf{Model Implementation.} We implement two parameterizations: a linear model $R_\theta(x) = w^\top \phi(x)$ with $\theta = w \in \mathbb{R}^d$ (high interpretability, $<1\mu s$ inference), and a small gradient-boosted tree ensemble for nonlinear interactions.


\noindent \textbf{Monotonicity Constraints.}
To align the reward function with domain knowledge, we enforce \textit{monotonicity constraints}: negative (travel time, waiting time, and travel-to-activity ratio do not increase reward) and positive (POI popularity, category diversity, and preference alignment do not decrease reward). The linear model enforces these directly during optimization; tree-based models use post-hoc validation with re-weighting of non-monotonic features.


\subsubsection{Reward-Guided Ensemble Selection}
\label{sec:ensemble_selection}
The reward model selects among planner candidates and serves as the training signal for downstream refinement. Rather than committing to a single planner, we treat the planners plus the reward model as an \emph{implicit ensemble}.
Concretely, given a trip context $u$, the \textsc{Plan} stage generates a set of feasible itineraries $\mathcal{X}(u) = \{x_1, \dots, x_K\}$ using different planning algorithms. We then evaluate each candidate using the learned reward model and select the itinerary with the highest predicted user satisfaction:
$x^\star = \arg\max_{x \in \mathcal{X}(u)} R_\theta(x)$.

Diversity arises from algorithmic heterogeneity rather than parameter perturbations, distinguishing this from standard ensembles. A common human-aligned reward score enables direct comparison across planners with different internal objectives. Selection is lightweight, fully on-device, and adds negligible overhead to the planning stage.

\subsection{ADAPT: On-Device Itinerary Refinement}
\label{sec:adapt}

\textsc{Plan} produces diverse feasible itineraries; Learn provides a reward model $R_\theta(x)$ over itinerary-level features. Two \emph{practical and operational} gaps remain: (i) seed planners are not optimized for the learned reward, and (ii) mobile devices impose strict runtime and memory limits. \textsc{Adapt} addresses both via feasibility-preserving local edits under a resource-aware compute budget, with an anytime guarantee that returns the best feasible itinerary found so far.

\subsubsection{Local Feasible Edit Operators}
\label{sec:local_edits}

\textsc{Adapt} explores a neighborhood $\mathcal{N}(x)$ of a feasible itinerary $x$, defined by atomic edit operators with a validation layer that guarantees feasibility. Each edit modifies a small portion of the itinerary (typically within a day), followed by local re-timing from the first affected position; proposals that violate constraints are discarded.

\noindent\textbf{Atomic Operators.} Let a visit be indexed by $(d,i)$, denoting position $i$ on day $d$. We define the following six local moves:
\squishlist
    \item \textbf{Drop$(d,i)$:} remove the visit at position $i$ on day $d$.
    \item \textbf{Insert$(d,p,i)$:} insert an unvisited POI $p$ at position $i$ on day $d$.
    \item \textbf{Replace$(d,i,p)$:} replace the visit $(d,i)$ with an unvisited POI $p$.
    \item \textbf{SwapAdj$(d,i)$:} swap visits at positions $i$ and $i{+}1$ on day $d$.
    \item \textbf{Relocate$(d,i,j)$:} move the visit from position $i$ to $j$ on day $d$.
    \item \textbf{MoveDay$(d,i,d',j)$:} move the visit $(d,i)$ to $(d',j)$.
\squishend

These operators form a compact move set that covers common user-driven edits (``I want to add this museum'', ``swap lunch and park'', ``remove this stop'') while enabling local search to correct structural defects such as excessive travel chaining, poor pacing, or category imbalance.

\noindent\textbf{Feasibility Layer and Incremental Re-timing.}
After each edit, we reconstruct the schedule for the affected suffix via a forward pass. For day visits $(v_1,\dots,v_m)$ where $v_j$ corresponds to POI $p_j$ with duration $dur_{p_j}$ and window $[open_{p_j},\, close_{p_j}]$. With current cursor time $t$ at the end
of $v_{j-1}$, we compute
$arr_j = t + T(p_{j-1}, p_j)$, $start_j = \max\{arr_j,\, open_{p_j}\}$, and
$end_j = start_j + dur_{p_j}$. We accept the schedule iff (i) $p_j$ is open at $start_j$, (ii) $(start_j, end_j)$ lies within the user's
availability interval, (iii) $end_j \le day\_start + B_{\text{schedule}}$ (the user's daily time budget),
and (iv) $end_j$ does not exceed the day boundary. We then set $t \leftarrow end_j$ and continue. This
procedure strictly preserves feasibility when it succeeds, provides natural waiting-time
accounting via $start_j - arr_j$, and enables \emph{incremental} evaluation because only a local
suffix must be recomputed.


\subsubsection{Reward-Guided Local Search}
\label{sec:reward_guided_refine}

We refine a seed itinerary $x_0$ by iteratively applying local edits to improve  $R_\theta(x)$. Feasibility is enforced by the re-timing layer, so every accepted itinerary is feasible. We use a simple hill-climbing core with stochastic neighborhood sampling; more complex acceptance rules (e.g., simulated annealing) are compatible but unnecessary under tight budgets.

\noindent\textbf{Neighborhood Sampling.} 
Enumerating all feasible neighbors is expensive with many unvisited POIs. Instead, we sample a bounded candidate set per iteration: (i) an operator type with fixed probabilities (favoring \texttt{SwapAdj}, \texttt{Relocate}, \texttt{Replace}); (ii) target indices uniformly from the current schedule; (iii) for \texttt{Insert}/\texttt{Replace}, candidate POIs from a filtered pool (budget tier, category interest, distance). Each candidate is re-timed incrementally; infeasible proposals are discarded.

\noindent\textbf{Best-improvement under a fixed per-iteration cap.} 
Over the sampled candidates $\tilde{\mathcal{N}}(x)$ we select $x^{*} = \arg\max_{y \in \tilde{\mathcal{N}}(x)} R_\theta(y)$ and accept it if $R_\theta(x^{*}) - R_\theta(x) \geq \delta$. The procedure is anytime: $x_{\text{best}}$ is tracked globally and returned if the time budget expires.


\subsubsection{Resource-Aware Budgeting}
\label{sec:resource_aware}

PLA must run fully offline with predictable latency. \textsc{Adapt} therefore uses a \emph{device-aware} budget that bounds refinement runtime. At runtime we query device signals: model class, core count, available memory, and low-power mode. We map these to a refinement time budget: $B_{\text{ms}} = \texttt{clip}(B_0 \cdot s_{\text{cpu}} \cdot s_{\text{mem}} \cdot s_{\text{power}},\; B_{\min},\; B_{\max})$, where $B_0$ is a nominal budget (200–500 ms) and scaling factors come from coarse bins over the device signals. Coarse binning avoids fragile device-specific calibration, and the anytime search ensures graceful degradation on resource-constrained devices.



\section{Evaluation}
\label{sec:evaluation}

\begin{figure*}[!htb]
\begin{minipage}{0.31\textwidth}
\centering
\includegraphics[width=\columnwidth]{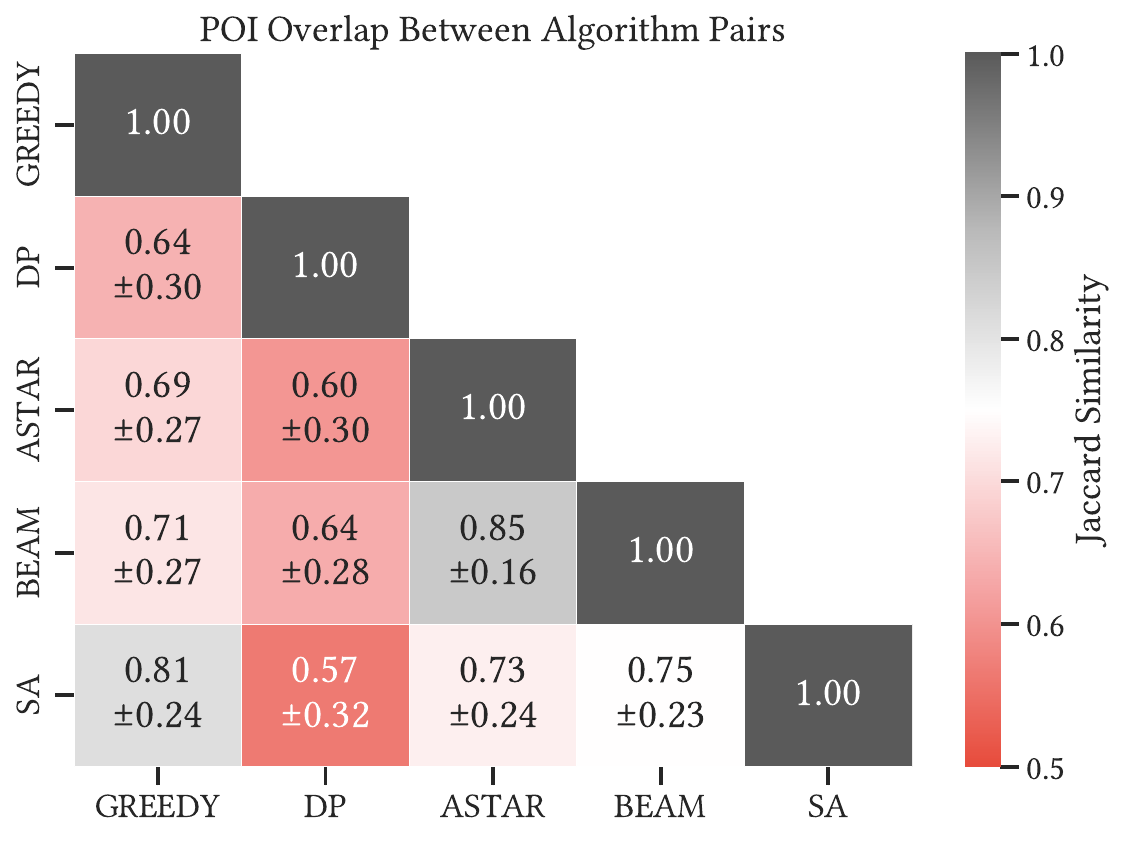}\vspace{-0.05in}
\caption{POI overlap (Jaccard similarity) between algorithm pairs. Mean overlap is 0.697; DP--SA show the most distinct selections.\vspace{-0.1in}}
\label{fig:poi_jaccard}
\end{minipage}
\hspace{.1em}
\begin{minipage}{0.31\textwidth}
\centering
\includegraphics[width=\columnwidth]{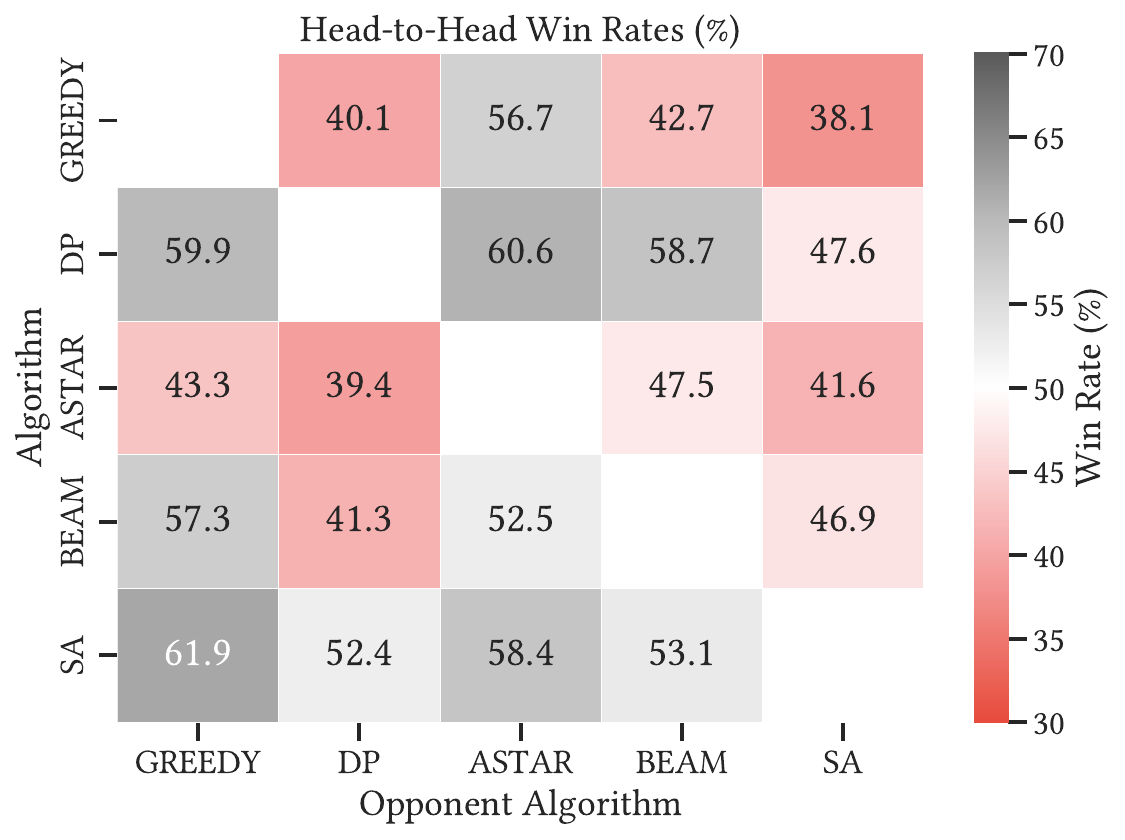}\vspace{-0.05in}
\caption{Head-to-head win rate matrix. No single algorithm dominates all matchups; the ten pairs exhibit complementary strengths.\vspace{-0.1in}}
\label{fig:h2h_matrix}
\end{minipage}
\hspace{.1em}
\begin{minipage}{0.31\textwidth}
\centering
\includegraphics[width=\columnwidth]{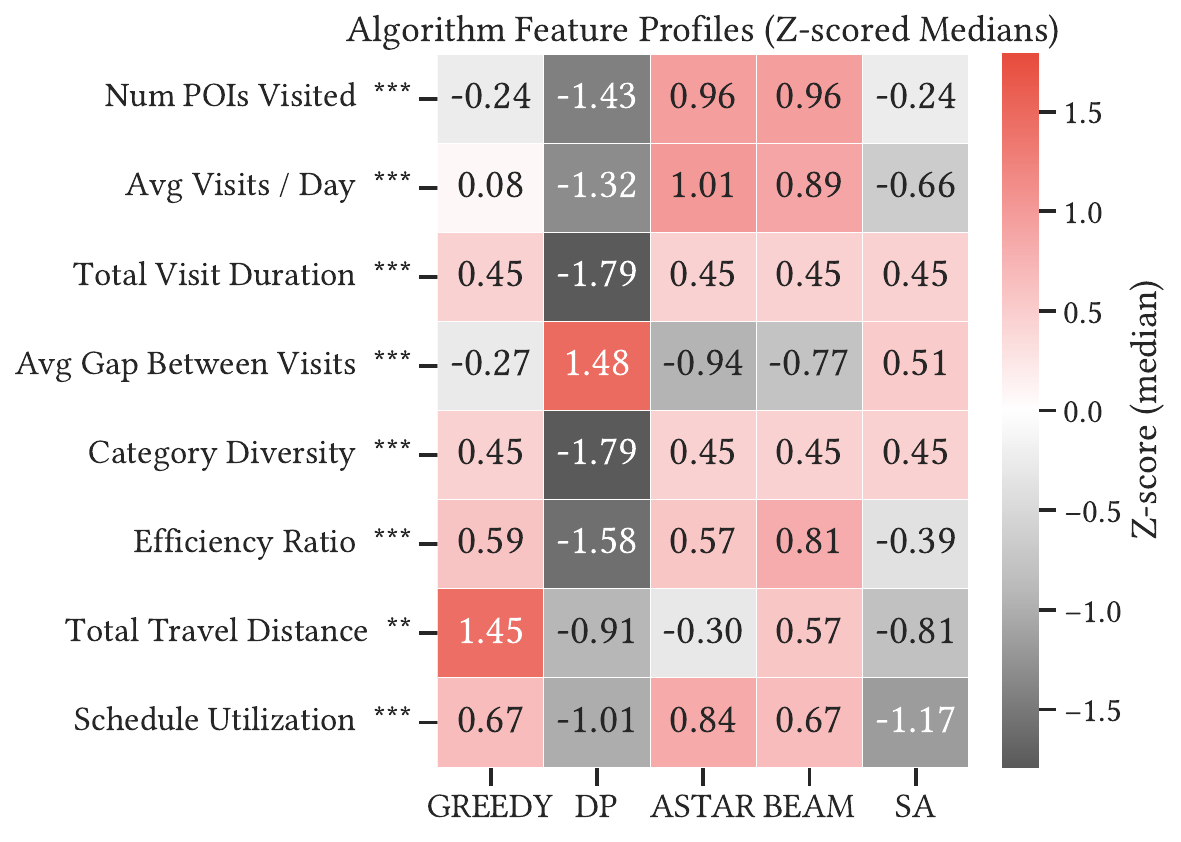}\vspace{-0.05in}
\caption{Z-scored median feature profiles per algorithm (deviation from cross-algorithm mean). All 8 features differ significantly ($p<0.05$).\vspace{-0.1in}}
\label{fig:feature_diversity}
\end{minipage}
\end{figure*}

We evaluate the PLA framework along five axes: (1)~[\textsc{Plan}] algorithmic diversity, (2)~[\textsc{Learn}] model accuracy, (3)~[\textsc{Plan}~+~\textsc{Learn}] ensemble effectiveness, (4)~[\textsc{Adapt}] refinement uplift, and (5)~[\textsc{Plan}~+~\\ \textsc{Learn}~+~\textsc{Adapt}] overall performance.

\subsection{Dataset and Experimental Setup}
\label{sec:eval_setup}
Our evaluation uses 2,519 pairwise comparisons across 100+ U.S. cities. Each comparison shows side-by-side itineraries from two randomly chosen planners (full protocol in Section~\ref{sec:preference_collection}). Coverage spans 5 planners, 1–7 day trips, all schedule-intensity × budget-tier combinations, and varied interest profiles. We report 5-fold stratified cross-validation results with bootstrap 95\% confidence intervals (1,000 resamples).



\noindent\textbf{Inter-Annotator Agreement.}
Since each comparison is assigned to one annotator, we ran a dedicated IAA study where two annotators independently evaluated 50 shared itinerary pairs. The study yields 78.0\% raw agreement (39/50), Cohen's $\kappa = 0.56$, Krippendorff's $\alpha = 0.56$, with balanced choices (27:23, 26:24). Across the full dataset we find no position bias (Plan 1 win rate 50.9\%, p = 0.38) and low annotator heterogeneity (std 0.10 across 17 annotators). The moderate $\kappa$ matches subjective preference tasks and motivates Bradley–Terry modeling, which treats disagreement as probabilistic signal. The 78.0\% human-human rate bounds attainable model accuracy: the deployed 66.5\% model (Section~\ref{sec:learn_evaluation}) captures ~60\% of the gap between random and human agreement.

\subsection{[\textsc{Plan}] Algorithmic Diversity}
\label{sec:plan_evaluation}
A key hypothesis of PLA is that heterogeneous planners produce structurally diverse feasible candidates. We validate this via POI overlap, head-to-head win rate, and feature diversity analyses.

\noindent\textbf{POI Overlap Analysis.}
We measure itinerary differences via Jaccard similarity of POI sets (Figure~\ref{fig:poi_jaccard}). Mean Jaccard similarity is 0.697 across algorithm pairs, indicating algorithms share approximately 70\% of POIs on average for the same trip context. DP--SA pairs are most distinct (0.565) and A*–Beam pairs most similar (0.847). Even so, 10.6\% of comparisons fall below 30\% POI overlap, indicating genuine exploration of different solution regions.



\begin{figure*}[!htb]
\begin{minipage}{0.31\textwidth}
\centering
\captionsetup{type=table}
\caption{Pairwise prediction (5-fold CV).\vspace{-0.1in}}
\label{tab:baselines}
\resizebox{\linewidth}{!}{
\begin{tabular}{lc}
\toprule
\textbf{Method} & \textbf{Accuracy} \\
\midrule
Random & 50.0\% \\
Majority Class & 51.0\% \\
Heuristic: \texttt{num\_pois} & 52.2\% \\
Heuristic: \texttt{avg\_visits\_per\_day} & 51.6\% \\
Heuristic: \texttt{total\_travel\_distance} & 46.6\% \\
\midrule
\textbf{Bradley-Terry Model (deployed)} & \textbf{66.5\%} \\
\bottomrule
\end{tabular}
}
\end{minipage}
\hspace{.1em}
\begin{minipage}{0.31\textwidth}
\centering
\captionsetup{type=table}
\caption{Model comparison (5-fold CV).\vspace{-0.1in}}
\label{tab:model_comparison}
\resizebox{\linewidth}{!}{
\begin{tabular}{lcc}
\toprule
\textbf{Model}  (Deployment Pref.$\downarrow$) & \textbf{ROC-AUC} & \textbf{F1} \\
\midrule
LR (L2) & 0.728 & 0.687 \\
SVM (RBF) & 0.723 & 0.696 \\
Random Forest & 0.755 & 0.707 \\
XGBoost & 0.751 & 0.706 \\
Ensemble (LR+XGBoost) & 0.752 & 0.696 \\
Ensemble (LR+RF+SVM) & 0.750 & 0.704 \\
\midrule
\textbf{LR (ElasticNet) [deployed]} & \textbf{0.730} & \textbf{0.687} \\
\bottomrule
\end{tabular}
}
\end{minipage}
\hspace{.1em}
\begin{minipage}{0.31\textwidth}
\centering
\captionsetup{type=table}

\caption{Counterfactual analysis.\vspace{-0.1in}}
\label{tab:ensemble}
\resizebox{\linewidth}{!}{
\begin{tabular}{lccc}
\toprule
\textbf{Strategy} & \textbf{Win Rate} & \textbf{95\% CI} & \textbf{n} \\
\midrule
Random Selection & 50.0\% & --- & --- \\
Greedy Only & 44.6\% & [41.5\%, 47.8\%] & 1024 \\
A* Only & 42.9\% & [39.7\%, 46.1\%] & 1005 \\
Beam Only & 49.6\% & [46.4\%, 52.8\%] & 987 \\
SA Only & 56.4\% & [53.4\%, 59.5\%] & 998 \\
DP Only & 56.6\% & [53.6\%, 59.7\%] & 1024 \\
\midrule
\textbf{Ensemble + $R_\theta$} & \textbf{67.8\%} & [65.9\%, 69.6\%] & 2519 \\
\bottomrule
\end{tabular}
}
\end{minipage}
\vspace{-0.5em}
\end{figure*}

\noindent\textbf{Win Rate Analysis.}
Figure~\ref{fig:h2h_matrix} shows the head-to-head win rate matrix across all 10 algorithm pairs. All pairs are complementary: both algorithms win in at least some comparisons. DP beats A* most decisively (60.6\%–39.4\%); DP–SA is the most competitive (52.4\%–47.6\%). Complementarity is a prerequisite for ensemble selection: if one algorithm dominated, selection would collapse to that algorithm.



\noindent\textbf{Feature Diversity Analysis.}
Beyond POI selection, we examine structural differences. Figure~\ref{fig:feature_diversity} shows Z-scored median feature profiles for each algorithm. A* and Beam produce the densest itineraries with the tightest visit spacing, DP the sparsest with the largest inter-visit gaps, while Greedy incurs the longest travel distances and SA the lowest schedule utilization. Kruskal-Wallis tests confirm \emph{all 8 features} differ significantly across algorithms ($p < 0.05$), including number of POIs, visits per day, travel distance, and category diversity. Planners occupy different regions of feature space, producing structurally different itineraries.



\subsection{[\textsc{Learn}] Model Accuracy}
\label{sec:learn_evaluation}
We evaluate the reward model's accuracy against heuristic baselines, compare architectures and input formulations, and characterize data requirements via learning curves.

\noindent\textbf{Heuristic Baseline Comparison.}
In Table~\ref{tab:baselines}, learned Bradley--Terry model achieves \textbf{66.5\% $\pm$ 2.3\%} accuracy (5-fold CV), outperforming all single-feature heuristics. The improvement over the best heuristic, (\texttt{num\_pois} at 52.2\%) is $+14.3\%$ absolute. The model captures feature interactions that no single metric can express.

\noindent\textbf{Model Architecture Comparison.}
Table~\ref{tab:model_comparison} compares architectures under the Bradley–Terry formulation (5-fold CV). The deployed Logistic Regression (ElasticNet) achieves ROC-AUC 0.730 and F1 0.687. Random Forest reaches ROC-AUC 0.755 and F1 0.707, but the ROC-AUC and F1 gaps are within bootstrap variability and less relevant for ensemble selection, which depends on within-context ranking rather than global pair discrimination (Section~\ref{sec:ensemble_selection}). We deploy the linear model: it offers per-feature interpretability for production diagnosis, native enforcement of the monotonicity constraints (Section~\ref{sec:reward_model}), and an order-of-magnitude smaller on-device footprint.



\begin{figure*}[!htb]
\begin{minipage}{0.31\textwidth}
\centering
\includegraphics[width=\columnwidth]{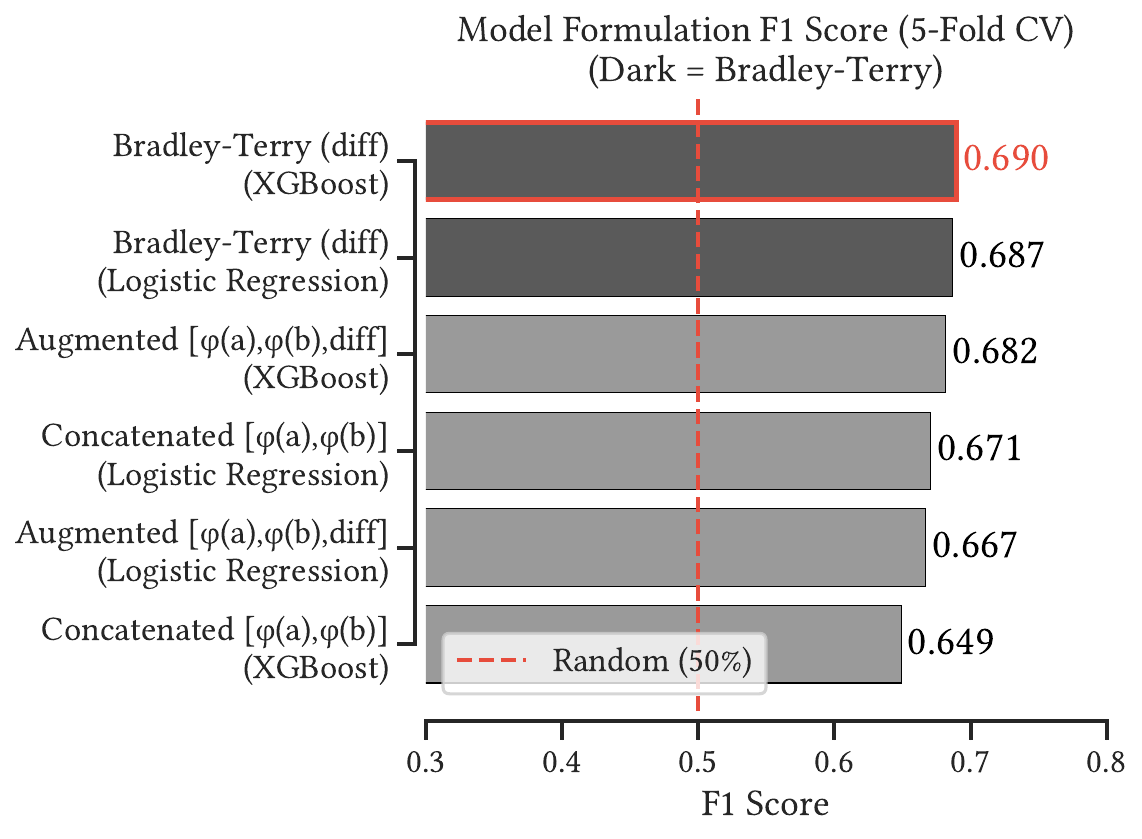}\vspace{-0.05in}
\caption{F1 score by input formulation. Bradley-Terry (difference-based, dark bars) outperforms concatenated and augmented alternatives.\vspace{-0.1in}}
\label{fig:formulation_f1}
\end{minipage}
\hspace{.1em}
\begin{minipage}{0.31\textwidth}
\centering
\includegraphics[width=\columnwidth]{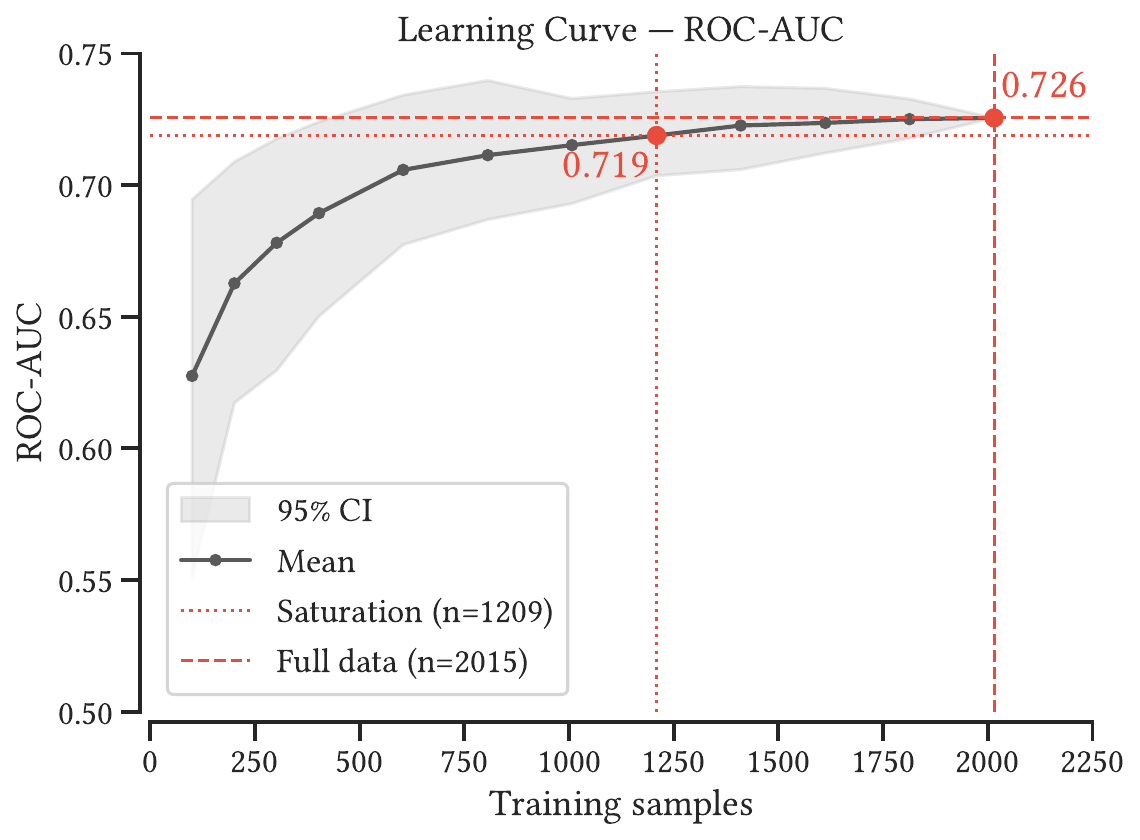}\vspace{-0.05in}
\caption{Learning curve (ROC-AUC vs.\ training size). Performance plateaus near 1,209 samples, indicating data sufficiency under the current feature set.\vspace{-0.1in}}
\label{fig:learning_curve}
\end{minipage}
\hspace{.1em}
\begin{minipage}{0.31\textwidth}
\centering
\includegraphics[width=\columnwidth]{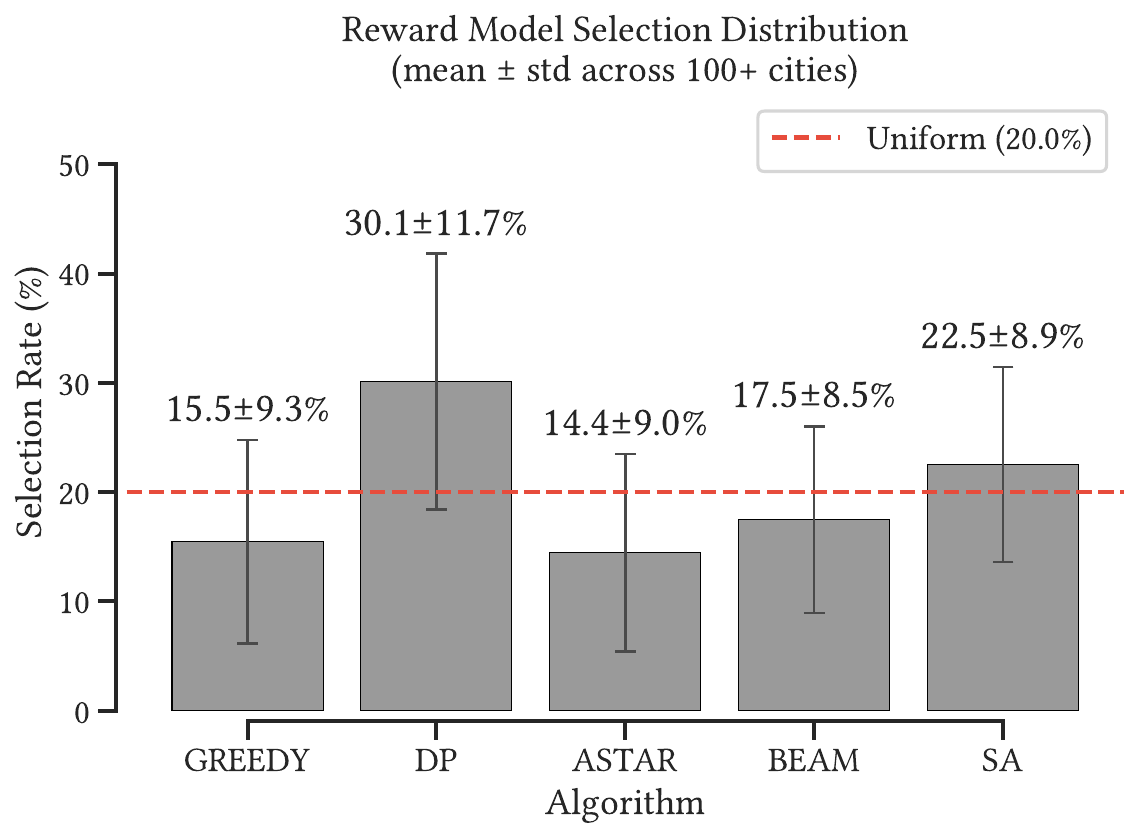}\vspace{-0.05in}
\caption{Algorithm selection distribution across cities (±std). Near-maximum entropy (97.6\%) indicates all algorithms contribute to the ensemble.
\vspace{-0.1in}}
\label{fig:selection_distribution}
\end{minipage}
\vspace{-0.5em}
\end{figure*}

\noindent\textbf{Model Formulation Comparison.}
Figure~\ref{fig:formulation_f1} shows F1 score by input formulation. The Bradley--Terry (difference-based) formulation achieves the highest F1, outperforming concatenated and augmented alternatives. The difference-based representation thus captures pairwise structure more effectively than formulations treating the two candidates independently.

\noindent\textbf{Learning Curve Analysis.}
Figure~\ref{fig:learning_curve} shows model ROC-AUC as training size grows. AUC improves from 0.628 ($n=100$) to 0.719 ($n = 1{,}209$); doubling to 2,015 adds only $+0.007$, within bootstrap CIs. The 2,519-comparison dataset is in the saturated regime; additional gains would require richer features, not more data. The feature set is deliberately constrained for three on-device requirements: sub-millisecond inference, native monotonicity enforcement (Section~\ref{sec:reward_model}), and per-feature interpretability. Richer representations (e.g., learned POI-sequence embeddings) are natural extensions when these constraints can be relaxed.
Leave-one-city-out (LOCO) cross-validation assesses cross-city generalization. The model achieves 67.6\% mean accuracy (8.6\% std) across 100+ held-out cities, with per-city accuracy ranging from 51.5\% to 90.0\%. The variation reflects per-city discrimination difficulty: cities with less diverse POI distributions are harder to rank than data-rich cities.

\subsection{[\textsc{Plan} + \textsc{Learn}] Ensemble Effectiveness}
\label{sec:ensemble_evaluation}

We evaluate whether reward-model-guided selection outperforms any single algorithm, examine the diversity of algorithm selection across contexts, and validate that predicted reward scores align with empirical human preferences.

\noindent\textbf{Ensemble Model Performance.}
Table~\ref{tab:ensemble} compares the reward-model ensemble against individual algorithms via counterfactual analysis on the full 2,519 comparisons. The ensemble achieves a \textbf{67.8\%} win rate, versus 56.6\% for the best single algorithm (DP), a gain of \textbf{$+11.2\%$} absolute. The 95\% bootstrap CI [65.9\%, 69.6\%] does not overlap with DP's interval [53.6\%, 59.7\%], confirming statistical significance (McNemar's test $p < 0.001$).

\noindent\textbf{Ensemble Selection Diversity.}
Figure~\ref{fig:selection_distribution} shows which algorithm is selected by the reward model. DP is selected most often (30.1\%), followed by SA (22.5\%), Beam (17.5\%), Greedy (15.5\%), and A* (14.4\%). The selection entropy is 2.265 bits out of a maximum 2.322 bits (97.6\% of maximum entropy), indicating highly diverse selection with no single algorithm dominating. The chi-square test confirms significant non-uniformity ($p < 0.001$), validating that different algorithms excel in different contexts.

\begin{figure*}[!htb]
\begin{minipage}{0.31\textwidth}
\centering
\includegraphics[width=\columnwidth]{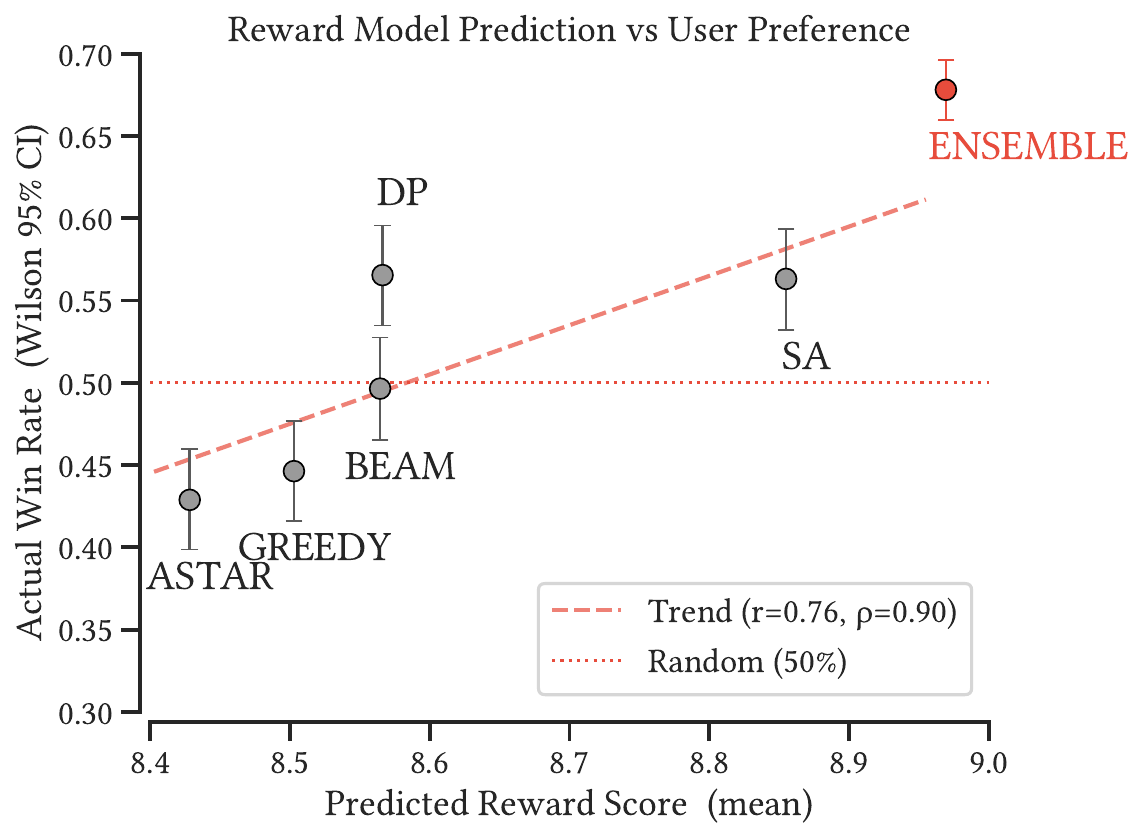}\vspace{-0.05in}
\caption{Predicted reward score vs.\ empirical win rate for algorithms. Correlation ($\rho = 0.90$) between predicted score and win rate shows that score maximization tracks human preference.\vspace{-0.1in}}
\label{fig:predicted_vs_actual}
\end{minipage}
\hspace{.1em}
\begin{minipage}{0.31\textwidth}
\centering
\includegraphics[width=\columnwidth]{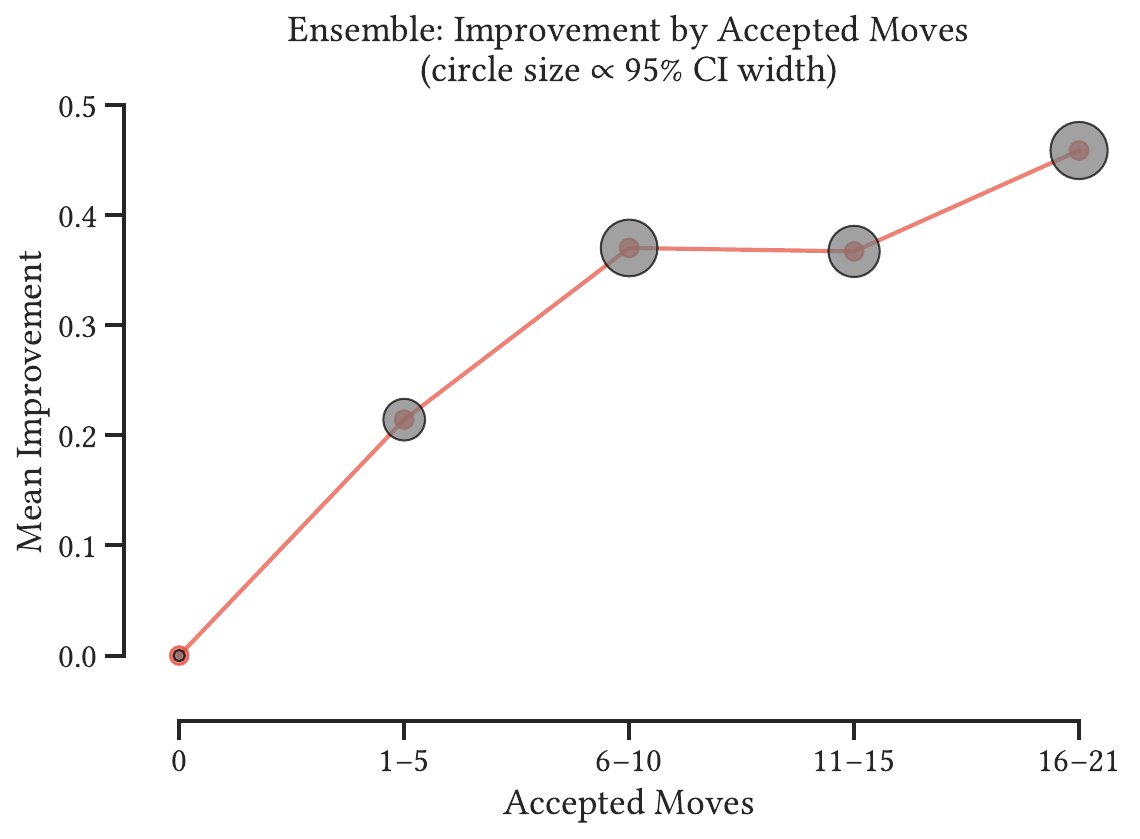}\vspace{-0.05in}
\caption{Ensemble improvement by number of accepted moves (binned). Each additional accepted move typically contributes to score gain, with higher variance at higher move counts.\vspace{-0.1in}} 
\label{fig:adapt_iter}
\end{minipage}
\hspace{.1em}
\begin{minipage}{0.31\textwidth}
\centering
\includegraphics[width=\columnwidth]{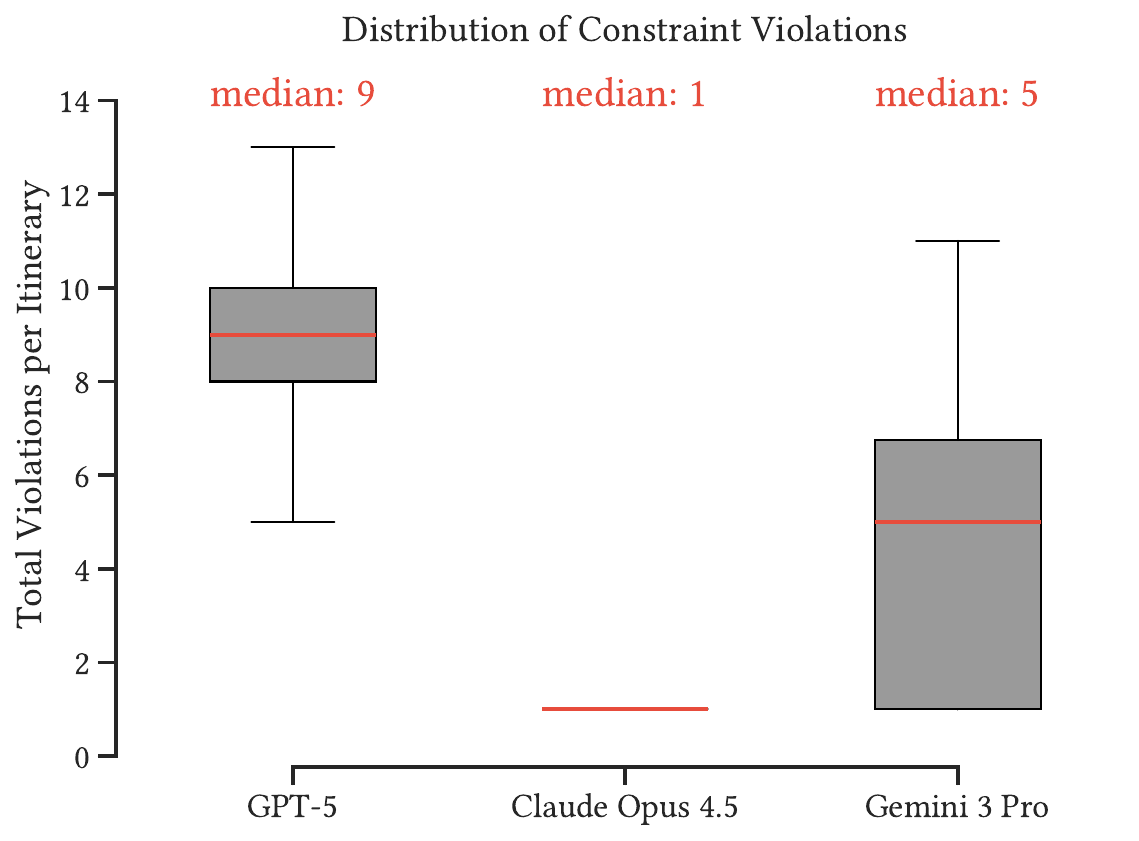}\vspace{-0.05in}
\caption{Distribution of constraint violations per itinerary for three frontier LLMs. All models achieve 0\% strict feasibility; whereas PLA maintains 100\% feasibility by construction.\vspace{-0.1in}}
\label{fig:llm_violations}
\end{minipage}
\vspace{0.5em}
\end{figure*}

\noindent\textbf{Ensemble Score Improvement.}
Figure~\ref{fig:predicted_vs_actual} plots each algorithm's mean predicted reward against its empirical win rate, with 95\% Wilson confidence interval error bars. The Spearman correlation between predicted score and empirical win rate is $\rho = 0.90$, so the reward model's ranking aligns with annotator preferences. The ensemble (shown in red) attains both a higher predicted score and a higher win rate than any individual algorithm; maximizing the reward model therefore yields itineraries that humans prefer.

\subsection{[\textsc{Adapt}] Refinement Uplift}
\label{sec:adapt_evaluation}

We quantify the score uplift delivered by local refinement across all seed algorithms and characterize how improvement scales with the number of accepted moves. Reward score improvement serves as a proxy for human preference improvement: Section~\ref{sec:ensemble_evaluation} shows that reward model scores predict human-labeled win rates at Spearman $\rho = 0.90$, indicating that score maximization reliably selects annotator-preferred itineraries.

\noindent\textbf{Refinement Score Improvement.}
Table~\ref{tab:adapt} shows reward before and after \textsc{Adapt} refinement. All improvements are statistically significant ($p < 0.001$, Wilcoxon signed-rank test). Algorithms with lower initial reward scores benefit more: A* gains $+11.7\%$, while the ensemble seed (already the highest-scoring before refinement) improves by $+3.9\%$. \textsc{Adapt} thus provides uplift regardless of seed quality, and ensemble selection followed by refinement produces the strongest final itinerary.

\noindent\textbf{Refinement Iteration Improvement.}
Figure~\ref{fig:adapt_iter} shows ensemble improvement grouped by number of accepted moves, binned in intervals of five. Itineraries with more accepted moves show larger score gains; each accepted move generally contributes to quality improvement. This upward trend is consistent with the hill-climbing acceptance criterion, which only incorporates score-improving proposals, though variance increases at higher moves.


\subsection{[\textsc{Plan} + \textsc{Learn} + \textsc{Adapt}] Performance}
\label{sec:e2e_evaluation}

Combining all three stages, we benchmark PLA against LLM-based generators on feasibility, report on-device latency, and quantify real-world deployment impact. We do not compare against prior itinerary systems (TravelPlanner, PersonalTravelSolver) because they target a different operating point: open-ended natural-language input, single-day or unbounded-budget formulations, and cloud inference. Adapting them to our multi-day, on-device regime would measure adaptation, not the methods. The classical optimization baselines are covered by the per-planner comparisons in Section~\ref{sec:ensemble_evaluation}, while the LLMs are evaluated in the following comparison.


\noindent\textbf{LLM Benchmark Comparison.}
We evaluated 294 itineraries (98 each from GPT-5, Claude Opus 4.5, Gemini 3 Pro) against our full hard-constraint set. Strict feasibility rate is \textbf{0/98 (0\%)} for all three models, with a Wilson 95\% CI upper bound of $\approx 3.8\%$ per model. Figure~\ref{fig:llm_violations} shows total violation distributions: Claude Opus 4.5 has the fewest (median 1, mean 1.08), then Gemini 3 Pro (5, 4.44) and GPT-5 (9, 9.03). Table~\ref{tab:llm_violation_breakdown} shows that travel-time consistency dominates (an average of 1.04, 3.28, and 5.98 violations per itinerary for Claude Opus~4.5, Gemini~3~Pro, and GPT-5), with visit-duration second (0.03, 0.22, and 1.62, respectively); other categories rarely violate. The constraint regime is hard for autoregressive generation. On the other hand, PLA offers guaranteed feasibility at interactive on-device latency, versus approximate feasibility from a cloud API.

\begin{figure*}[!htb]
\begin{minipage}{0.31\textwidth}
\centering
\captionsetup{type=table}
\caption{\textsc{Adapt} refinement by seed. All score improvements are statistically significant ($p < 0.001$).\vspace{-0.1in}}
\label{tab:adapt}
\resizebox{\linewidth}{!}{
\begin{tabular}{lcccc}
\toprule
\textbf{Seed} & \textbf{Before} & \textbf{After} & \textbf{Uplift} & \textbf{95\% CI} \\
\midrule
Greedy & 6.835 & 7.524 & +10.1\% & [0.563, 0.825] \\
DP     & 6.955 & 7.268 & +4.5\%  & [0.213, 0.431] \\
A*     & 6.623 & 7.399 & +11.7\% & [0.671, 0.883] \\
Beam   & 6.805 & 7.390 & +8.6\%  & [0.501, 0.676] \\
SA     & 6.983 & 7.491 & +7.2\%  & [0.425, 0.597] \\
\midrule
\textbf{Ensemble} & \textbf{7.184} & \textbf{7.463} & \textbf{+3.9\%} & [0.191, 0.392] \\
\bottomrule
\end{tabular}
}

\end{minipage}
\hspace{.1em}
\begin{minipage}{0.31\textwidth}
\centering
\captionsetup{type=table}
\caption{Mean constraint violations per itinerary by type (mean $\pm$ std) for GPT-5, Claude Opus 4.5, Gemini 3 Pro.\vspace{-0.1in}}
\label{tab:llm_violation_breakdown}
\resizebox{\linewidth}{!}{
\begin{tabular}{lccc}
\toprule
\textbf{Violation} & \textbf{GPT-5} & \textbf{Opus 4.5} & \textbf{Gemini 3} \\
\midrule
Travel Time  & $5.98 \pm 1.51$ & $1.04 \pm 0.32$ & $3.28 \pm 1.85$ \\
Visit Hours  & $1.36 \pm 0.84$ & $0.01 \pm 0.10$ & $0.79 \pm 0.83$ \\
Duration     & $1.62 \pm 1.84$ & $0.03 \pm 0.17$ & $0.22 \pm 0.49$ \\
Time Budget  & $0.00 \pm 0.00$ & $0.00 \pm 0.00$ & $0.13 \pm 0.40$ \\
No Repeats   & $0.05 \pm 0.22$ & $0.00 \pm 0.00$ & $0.00 \pm 0.00$ \\
\midrule
\textbf{Total} & $\mathbf{9.03 \pm 3.36}$ & $\mathbf{1.08 \pm 0.37}$ & $\mathbf{4.44 \pm 2.91}$ \\
\bottomrule
\end{tabular}
}

\end{minipage}
\hspace{.1em}
\begin{minipage}{0.31\textwidth}
\centering
\captionsetup{type=table}
\caption{Post-deployment engagement changes in \textsc{FlyEnJoy} relative to the single-heuristic baseline.\vspace{-0.1in}}
\label{tab:post_deployment}
\scriptsize
\begin{tabular}{llc}
\toprule
\textbf{Metric} & \textbf{Type} & \textbf{Change} \\
\midrule
Itinerary Initiations     & Usage      & $+83\%$   \\
Completion Rate           & Usage & $+91\%$   \\
\midrule
Feature DAU               & Engagement & $+22\%$   \\
DAU/WAU Ratio             & Engagement  & $+20$ pp  \\
\midrule
Time per Session          & Efficiency & $-8\%$    \\
Time per User             & Efficiency & $-38\%$   \\
\bottomrule
\end{tabular}

\end{minipage}
\vspace{-0.5em}
\end{figure*}


\noindent\textbf{On-Device Latency Analysis.}
To empirically validate viability, we measured end-to-end latency on 1,728 runs: 6 cities $\times$ 4 trip configurations $\times$ 24 algorithm configurations (Greedy/Beam Search, parallel/sequential fetching, Basic/Advanced/ML filtering, with/without optimizations) $\times$ 3 repetitions. Latency averages 109.9 ms (std: 69.5, p5 35.1, p95 253.1 ms). Both the mean and the p95 tail fall below the 300 ms interactive-response threshold. The device-aware budget bounds worst-case latency across heterogeneous devices.


\noindent\textbf{Post-Deployment Quantification.}
We deployed PLA in \textsc{FlyEnJoy}, a production iOS travel app. We evaluated real-world impact via anonymized logs in a pre-post observational analysis against the prior single-heuristic baseline; confounders (e.g., organic growth) cannot be fully excluded. Due to business sensitivity, we report only relative changes (Table~\ref{tab:post_deployment}). Itinerary initiations rose $83.33\%$, and completion rate $90.91\%$. Engagement also improved: itinerary-feature DAU rose $+22\%$ and DAU/WAU $+20$ pp (more repeat usage). Time per session ($-8\%$) and per user ($-38\%$) fell, reflecting fewer interactions per completed itinerary. Consistent gains across initiation, completion, repeat usage, and session efficiency suggest PLA reduces friction in addition to improving feasibility and preference alignment.


\section{Related Work}
\textbf{Itinerary Planning via Optimization and Learning.}
Classical itinerary planning uses combinatorial optimization to balance POI utility under time-budget constraints~\cite{lim2015personalized,chen2013automatic}. Extensions add real-world constraints~\cite{zhang2016trip} or learned preference models~\cite{wang2019empowering}. Reinforcement-learning approaches treat itinerary generation as sequential decision-making~\cite{chen2020itinerary,chen2022automatic,chen2023trip}. Optimization ensures feasibility, learning captures preferences; neither alone balances both. We separate feasibility-preserving planning from itinerary-level preference learning.

\noindent\textbf{Graph-Based, Multimodal, and LLM-Based Recommendation.}
Graph and sequence models address next-POI recommendation~\cite{sun2020go,yang2022getnext,qin2023disenpoi,yan2023spatio}, with multimodal extensions using geo-tagged images~\cite{juan2024multimodal}. LLM-based generators~\cite{li2024large,volchek2024chatgpt} use step-wise prediction or unconstrained generation and rarely guarantee global feasibility under temporal and resource constraints. Our work targets multi-day generation with feasibility guarantees and on-device execution.

\noindent\textbf{Hybrid Preference Learning and Constraint-Aware Planning.}
Hybrid approaches combine preference learning with constraint reasoning. Constraint-based recommenders enforce hard requirements with manually specified preferences~\cite{felfernig2015constraint}, while preference models often assume unconstrained candidates~\cite{burke2002hybrid}. Recent hybrid itinerary systems integrate preferences into optimization~\cite{tsai2025persire}. We differ by learning preferences at the itinerary level from pairwise judgments and optimizing them via feasibility-preserving refinement under a device-aware budget.

\section{Conclusion}
\label{sec:conclusion}

We presented PLA, a three-stage framework for personalized on-device itinerary generation. PLA composes complementary stages to address feasibility, desirability, and viability jointly. Across more than one hundred trip contexts, the planner ensemble reaches 97.6\% normalized selection entropy, meaning every algorithm contributes and none dominates. Itinerary-level quality such as pacing, day balance, and geographic coherence emerges from the learned reward model rather than from any per-POI score, and feasibility-preserving refinement removes the need for the post-hoc repair stage typical of LLM pipelines. In production deployment within \textsc{FlyEnJoy}, PLA increased itinerary completion rates by 91\%, providing a blueprint for human-aligned planning under real deployment constraints.

\section*{GenAI Usage Disclosure}
During the preparation of this work, the authors used LLM-based tools to assist with condensing and improving the clarity of the author-written text within selected sections. No new content, results, analyses or claims were created by AI tools. All AI-assisted edits were reviewed, verified and approved by the authors. The authors take full responsibility for the accuracy and integrity of all content in this paper.

\balance
\bibliographystyle{ACM-Reference-Format}
\bibliography{references,Itinerary}


\end{document}